# A Pattern Matching Method for Finding Noun and Proper Noun Translations from Noisy Parallel Corpora

Pascale Fung
Computer Science Department
Columbia University
New York, NY 10027
pascale@cs.columbia.edu

## Abstract

We present a pattern matching method for compiling a bilingual lexicon of nouns and proper nouns from unaligned, noisy parallel texts of Asian/Indo-European language pairs. Tagging information of one language is used. Word frequency and position information for high and low frequency words are represented in two different vector forms for pattern matching. New anchor point finding and noise elimination techniques are introduced. We obtained a 73.1% precision. We also show how the results can be used in the compilation of domain-specific noun phrases.

## 1 Bilingual lexicon compilation without sentence alignment

Automatically compiling a bilingual lexicon of nouns and proper nouns can contribute significantly to breaking the bottleneck in machine translation and machine-aided translation systems. Domain-specific terms are hard to translate because they often do not appear in dictionaries. Since most of these terms are nouns, proper nouns or noun phrases, compiling a bilingual lexicon of these word groups is an important first step.

We have been studying robust lexicon compilation methods which do not rely on sentence alignment. Existing lexicon compilation methods (Kupiec 1993; Smadja & McKeown 1994; Kumano & Hirakawa 1994; Dagan *et al.* 1993; Wu & Xia 1994) all attempt to extract pairs of words or compounds that are translations of each other from previously sentence-aligned, parallel texts. However, sentence alignment (Brown *et al.* 1991; Kay & Röscheisen 1993; Gale & Church 1993; Church 1993; Chen 1993; Wu 1994) is not always practical when corpora have unclear sentence boundaries or with noisy text segments present in only one language.

Our proposed algorithm for bilingual lexicon acquisition bootstraps off of corpus alignment procedures we developed earlier (Fung & Church 1994; Fung & McKeown 1994). Those procedures attempted to align texts by finding matching word pairs and have demonstrated their effectiveness for Chinese/English and Japanese/English. The main focus then was accurate alignment, but the procedure produced a small number of word translations as a by-product. In contrast, our new algorithm performs a minimal alignment, to facilitate compiling a much larger bilingual lexicon.

The paradigm for Fung & Church (1994); Fung & McKeown (1994) is based on two main steps - find a small bilingual *primary lexicon*, use the text segments which contain some of the word pairs in the lexicon as anchor points for alignment, align the text, and compute a better *secondary lexicon* from these partially aligned texts. This paradigm can be seen as analogous to the Estimation-Maximization step in Brown *et al.* (1991); Dagan *et al.* (1993); Wu & Xia (1994).

For a noisy corpus without sentence boundaries, the primary lexicon accuracy depends on the robustness of the algorithm for finding word translations given no *a priori* information. The reliability of the anchor points will determine the accuracy of the secondary lexicon. We also want an algorithm that bypasses a long, tedious sentence or text alignment step.

## 2 Algorithm overview

We treat the bilingual lexicon compilation problem as a pattern matching problem - each word shares some common features with its counterpart in the translated text. We try to find the best representations of these features and the best ways to match them. We ran the algorithm on a small Chinese/English parallel corpus of approximately 5760 unique English words.

The outline of the algorithm is as follows:

1. **Tag the English half of the parallel text.** In the first stage of the algorithm, only English words which are tagged as nouns or proper nouns are used to match words in the Chinese text.

2. **Compute the positional difference vector of each word**. Each of these nouns or proper nouns is converted from their positions in the text into a vector.

3. **Match pairs of positional difference vectors, giving scores**. All vectors from English and Chinese are matched against each other by Dynamic Time Warping (DTW).

4. **Select a primary lexicon using the scores**. A threshold is applied to the DTW score of each pair, selecting the most correlated pairs as the first bilingual lexicon.

5. **Find anchor points using the primary lexicon**. The algorithm reconstructs the DTW paths of these positional vector pairs, giving us a set of word position points which are filtered to yield anchor points. These anchor points are used for compiling a secondary lexicon.

6. **Compute a position binary vector for each word using the anchor points**. The remaining nouns and proper nouns in English and all words in Chinese are represented in a non-linear segment binary vector form from their positions in the text.

7. **Match binary vectors to yield a secondary lexicon**. These vectors are matched against each other by mutual information. A confidence score is used to threshold these pairs. We obtain the secondary bilingual lexicon from this stage.

In Section 3, we describe the first four stages in our algorithm, cumulating in a primary lexicon. Section 4 describes the next anchor point finding stage. Section 5 contains the procedure for compiling the secondary lexicon.

## 3 Finding high frequency bilingual word pairs

When the sentence alignments for the corpus are unknown, standard techniques for extracting bilingual lexicons cannot apply. To make matters worse, the corpus might contain chunks of texts which appear in one language but not in its translation[1], suggesting a discontinuous mapping between some parallel texts.

We have previously shown that using a vector representation of the frequency and positional information of a high frequency word was an effective way to match it to its translation (Fung & McKeown 1994). Dynamic Time Warping, a pattern recognition technique, was proposed as a good way to match these vectors. In our new algorithm, we use a similar positional difference vector representation and DTW matching techniques. However, we improve on the matching efficiency by installing tagging and statistical filters. In addition, we not only obtain a score from the DTW matching between pairs of words, but we also reconstruct the DTW paths to get the points of the best paths as anchor points for use in later stages.

### 3.1 Tagging to identify nouns

Since the positional difference vector representation relies on the fact that words which are similar in meaning appear fairly consistently in a parallel text, this representation is best for nouns or proper nouns because these are the kind of words which have consistent translations over the entire text.

As ultimately we will be interested in finding domain-specific terms, we can concentrate our effort on those words which are nouns or proper nouns first. For this purpose, we tagged the English part of the corpus by a modified POS tagger, and apply our algorithm to find the translations for words which are tagged as nouns, plural nouns or proper nouns only. This produced a more useful list of lexicon and again improved the speed of our program.

### 3.2 Positional difference vectors

According to our previous findings (Fung & McKeown 1994), a word and its translated counterpart usually have some correspondence in their frequency and positions although this correspondence might not be linear. Given the position vector of a word $p[i]$ where the values of this vector are the positions at which this word occurs in the corpus, one can compute a positional difference vector $V[i-1]$ where $V[i-1] = p[i] - p[i-1]$. $\dim(V)$ is the dimension of the vector which corresponds to the occurrence count of the word.

For example, if positional difference vectors for the word *Governor* and its translation in Chinese 總督 are plotted against their positions in the text, they give characteristic signals such as shown in Figure 1. The two vectors have different dimensions because they occur with different frequencies. Note that the two signals are shifted and warped versions of each other with some minor noise.

### 3.3 Matching positional difference vectors

The positional vectors have different lengths which complicates the matching process. Dynamic Time Warping was found to be a good way to match word vectors of shifted or warped forms (Fung & McKeown 1994). However, our previous algorithm only used the DTW score for finding the most correlated word pairs. Our new algorithm takes it one step further by backtracking to reconstruct the DTW paths and then automatically choosing the best points on these DTW paths as anchor points.

---

[1] This was found to be the case in the Japanese translation of the AWK manual (Church *et al.* 1993). The Japanese AWK was also found to contain different programming examples from the English version.

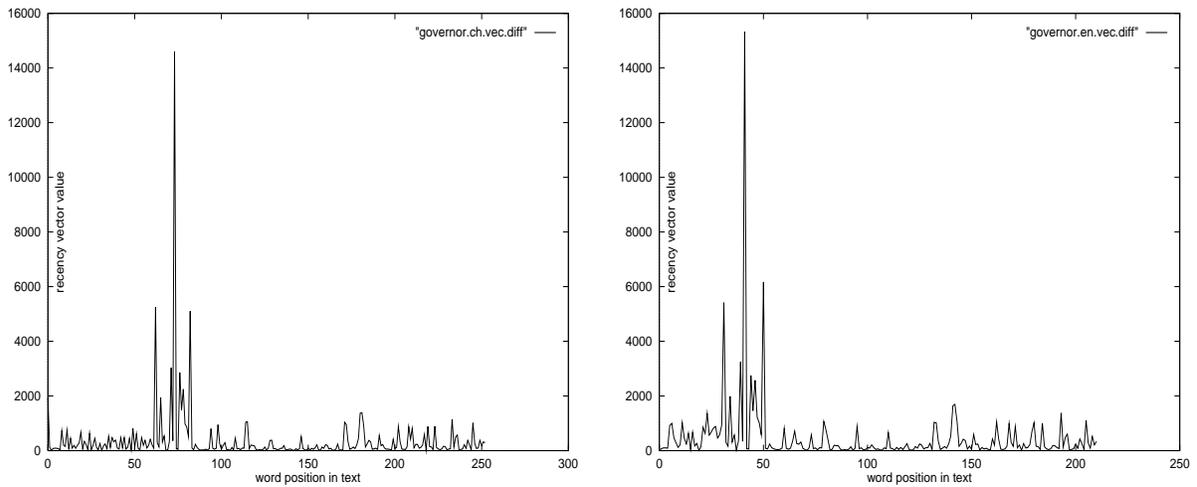

Figure 1: Positional difference signals showing similarity between *Governor* in English and Chinese

For a given pair of vectors $V1$, $V2$, we attempt to discover which point in $V1$ corresponds to which point in $V2$. If the two were not scaled, then position $i$ in $V1$ would correspond to position $j$ in $V2$ where $j/i$ is a constant. If we plot $V1$ against $V2$, we can get a diagonal line with slope $j/i$. If they occurred the same number of times, then every position $i$ in $V1$ would correspond to one and only one position $j$ in $V2$. For non-identical vectors, DTW traces the correspondences between all points in $V1$ and $V2$ (with no penalty for deletions or insertions). Our DTW algorithm with path reconstruction is as follows:

- **Initialization**

$$\begin{aligned}
\varphi_1(1,1) &= \zeta(1,1) \\
\varphi_1(i,1) &= \zeta(i,1) + \varphi(i-1,1]) \\
\varphi_1(1,j) &= \zeta(1,j) + \varphi(1,j-1) \\
\text{where } \varphi(a,b) &= \text{minimum cost of moving} \\
& \quad \text{from } a \text{ to } b \\
\zeta(c,d) &= |V1[c] - V2[d]| \\
\text{for } i &= 1, 2, \ldots, N \\
j &= 1, 2, \ldots, M \\
N &= \dim(V1) \\
M &= \dim(V2)
\end{aligned}$$

- **Recursion**

$$\begin{aligned}
\varphi_{n+1}(i,m) &= \min_{1 \leq l \leq 3}[\zeta(l,m) + \varphi_n(i,l)] \\
\xi_{n+1}(m) &= \operatorname*{argmin}_{1 \leq l \leq 3}[\zeta(l,m) + \varphi_n(i,l)] \\
\text{for } n &= 1, 2, \ldots, N-2 \\
\text{and } m &= 1, 2, \ldots, M
\end{aligned}$$

- **Termination**

$$\begin{aligned}
\varphi_N(i,j) &= \min_{1 \leq l \leq 3}[\zeta(l,m) + \varphi_{N-1}(i,l)] \\
\xi_N(j) &= \operatorname*{argmin}_{1 \leq l \leq 3}[\zeta(l,m) + \varphi_{N-1}(i,j)]
\end{aligned}$$

- **Path reconstruction**

  In our algorithm, we reconstruct the DTW path and obtain the points on the path for later use. The DTW path for *Governor*/總督 is as shown in Figure 2.

$$\begin{aligned}
\text{optimal path} &= (i, i_1, i_2, \ldots, i_{m-2}, j) \\
\text{where } i_n &= \zeta_{n+1}(i_{n+1}), \\
& \quad n = N-1, N-2, \ldots, 1 \\
\text{with } i_N &= j
\end{aligned}$$

We thresholded the bilingual word pairs obtained from above stages in the algorithm and stored the more reliable pairs as our primary bilingual lexicon.

### 3.4 Statistical filters

If we have to exhaustively match all nouns and proper nouns against all Chinese words, the matching will be very expensive since it involves computing all possible paths between two vectors, and then backtracking to find the optimal path, and doing this for all English/Chinese word pairs in the texts. The complexity of DTW is $\Theta(NM)$ and the complexity of the matching is $\Theta(IJNM)$ where $I$ is the number of nouns and proper nouns in the English text, $J$ is the number of unique words in the Chinese text, $N$ is the occurrence count of one English word and $M$ the occurrence count of one Chinese word.

We previously used some frequency difference constraints and starting point constraints (Fung & McKeown 1994). Those constraints limited the number of the pairs of vectors to be compared by DTW. For example, low frequency words are not considered since their positional difference vectors would not contain much information. We also apply these constraints in our experiments. However, there is still many pairs of words left to be compared.

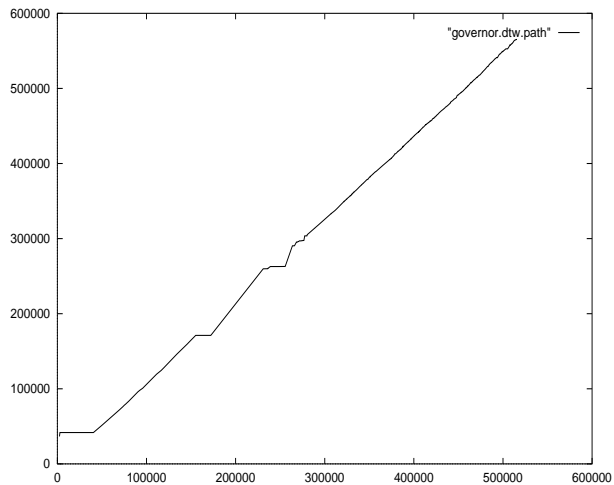

Figure 2: Dynamic Time Warping path for *Governor* in English and Chinese

To improve the computation speed, we constrain the vector pairs further by looking at the Euclidean distance $\mathcal{E}$ of their means and standard deviations:

$$\mathcal{E} = \sqrt{(m_1 - m_2)^2 + (\sigma_1 - \sigma_2)^2}$$

If their Euclidean distance is higher than a certain threshold, we filter the pair out and do not use DTW matching on them. This process eliminated most word pairs. Note that this Euclidean distance function helps to filter out word pairs which are very different from each other, but it is not discriminative enough to pick out the best translation of a word. So for word pairs whose Euclidean distance is below the threshold, we still need to use DTW matching to find the best translation. However, this Euclidean distance filtering greatly improved the speed of this stage of bilingual lexicon compilation.

## 4 Finding anchor points and eliminating noise

Since the primary lexicon after thresholding is relatively small, we would like to compute a secondary lexicon including some words which were not found by DTW. At stage 5 of our algorithm, we try to find anchor points on the DTW paths which divide the texts into multiple aligned segments for compiling the secondary lexicon. We believe these anchor points are more reliable than those obtained by tracing all the words in the texts.

For every word pair from this lexicon, we had obtained a DTW score and a DTW path. If we plot the points on the DTW paths of all word pairs from the lexicon, we get a graph as in the left hand side of Figure 3. Each point $(i, j)$ on this graph is on the DTW $path(v1, v2)$ where $v1$ is from English words in the lexicon and $v2$ is from the Chinese words in the lexicon. The union effect of all these DTW paths shows a salient line approximating the diagonal. This line can be thought of the text alignment path. Its departure from the diagonal illustrates that the texts of this corpus are not identical nor linearly aligned.

Since the lexicon we computed was not perfect, we get some noise in this graph. Previous alignment methods we used such as Church (1993); Fung & Church (1994); Fung & McKeown (1994) would bin the anchor points into continuous blocks for a rough alignment. This would have a smoothing effect. However, we later found that these blocks of anchor points are not precise enough for our Chinese/English corpus. We found that it is more advantageous to increase the overall reliability of anchor points by keeping the highly reliable points and discarding the rest.

From all the points on the union of the DTW paths, we filter out the points by the following conditions: If the point $(i, j)$ satisfies

| | |
|---|---|
| (slope constraint) | $j/i > 600 * N[0]$ |
| (window size constraint) | $i >= 25 + i_{previous}$ |
| (continuity constraint) | $j >= j_{previous}$ |
| (offset constraint) | $j - j_{previous} > 500$ |

then the point $(i, j)$ is noise and is discarded.

After filtering, we get points such as shown in the right hand side of Figure 3. There are 388 highly reliable anchor points. They divide the texts into 388 segments. The total length of the texts is around 100000, so each segment has an average window size of 257 words which is considerably longer than a sentence length; thus this is a much rougher alignment than sentence alignment, but nonetheless we still get a bilingual lexicon out of it.

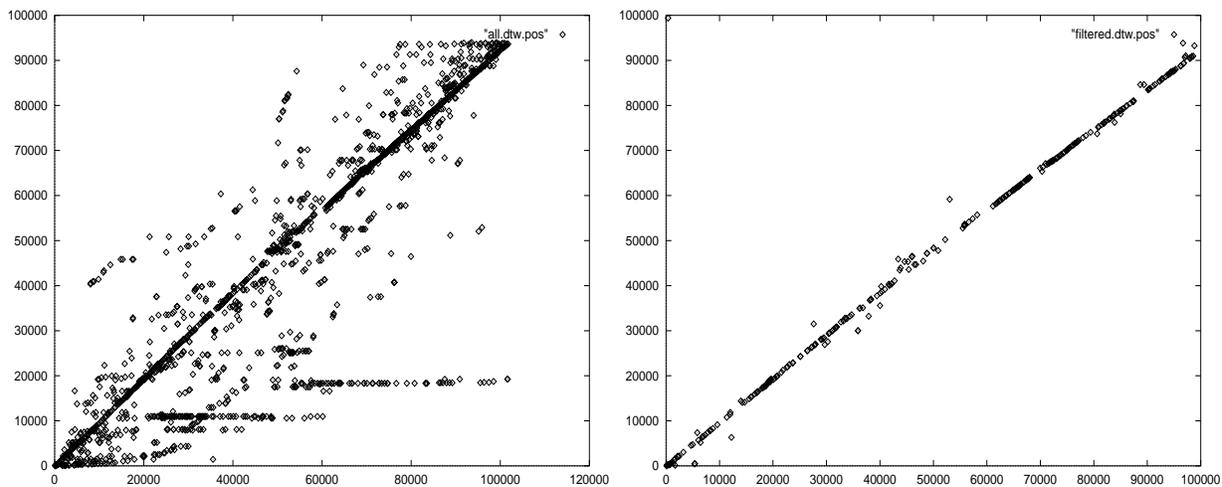

Figure 3: DTW path reconstruction output and the anchor points obtained after filtering

The constants in the above conditions are chosen roughly in proportion to the corpus size so that the filtered picture looks close to a clean, diagonal line. This ensures that our development stage is still unsupervised. We would like to emphasize that if they were chosen by looking at the lexicon output as would be in a supervised training scenario, then one should evaluate the output on an independent test corpus.

Note that if one chunk of noisy data appeared in text1 but not in text2, this part would be segmented between two anchor points $(i,j)$ and $(u,v)$. We know point $i$ is matched to point $j$, and point $u$ to point $v$, the texts between these two points are matched but we do not make any assumption about how this segment of texts are matched. In the extreme case where $i = u$, we know that the text between j and v is noise. We have at this point a segment-aligned parallel corpus with noise elimination.

## 5 Finding low frequency bilingual word pairs

Many nouns and proper nouns were not translated in the previous stages of our algorithm. They were not in the first lexicon because their frequencies were too low to be well represented by positional difference vectors.

### 5.1 Non-linear segment binary vectors

In stage 6, we represent the positional and frequency information of low frequency words by a binary vector for fast matching.

The 388 anchor points $(95, 10)$, $(139, 131)$, ..., $(98809, 93251)$ divide the two texts into 388 non-linear segments. Text1 is segmented by the points $(95, 139, \ldots, 98586, 98809)$ and text2 is segmented by the points $(10, 131, \ldots, 90957, 93251)$.

For the nouns we are interested in finding the translations for, we again look at the position vectors. For example, the word *prosperity* occurred seven times in the English text. Its position vector is $\langle 2178, 5322, \ldots, 86521, 95341 \rangle$. We convert this position vector into a binary vector $V1$ of 388 dimensions where $V1[i] = 1$ if *prosperity* occured within the $i$th segment, $V1[i] = 0$ otherwise. For *prosperity*, $V1[i] = 1$ where $i = 20, 27, 41, 47, 193, 321, 360$. The Chinese translation for prosperity is 繁榮. Its position vector is $\langle 1955, 5050, \ldots, 88048 \rangle$. Its binary vector is $V2[i] = 1$ where $i = 14, 29, 41, 47, 193, 275, 321, 360$. We can see that these two vectors share five segments in common.

We compute the segment vector for all English nouns and proper nouns not found in the first lexicon and whose frequency is above two. Words occurring only once are extremely hard to translate although our algorithm was able to find some pairs which occurred only once.

### 5.2 Binary vector correlation measure

To match these binary vectors $V1$ with their counterparts in Chinese $V2$, we use a mutual information score $m$.

$$m = \log_2 \frac{\Pr(V1, V2)}{\Pr(V1)\Pr(V2)}$$

$$\Pr(V1) = \frac{\text{freq}(V1[i] = 1)}{L}$$

$$\Pr(V2) = \frac{\text{freq}(V2[i] = 1)}{L}$$

$$\Pr(V1, V2) = \frac{\text{freq}(V1[i] = V2[i] = 1)}{L}$$

$$\text{where} \quad L = \dim(V1) = \dim(V2)$$

If *prosperity* and 繁榮 occurred in the same eight segments, their mutual information score would be 5.6. If they never occur in the same segments, their $m$ would be negative infinity. Here, for *prosperity*/繁榮, $m = 5.077$ which shows that these two words are indeed highly correlated.

The $t$-score was used as a confidence measure. We keep pairs of words if their $t > 1.65$ where

$$t \approx \frac{\Pr(V1, V2) - \Pr(V1)\Pr(V2)}{\sqrt{\frac{1}{L}\Pr(V1, V2)}}$$

For *prosperity*/繁榮, $t = 2.33$ which shows that their correlation is reliable.

## 6  Results

The English half of the corpus has 5760 unique words containing 2779 nouns and proper nouns. Most of these words occurred only once. We carried out two sets of evaluations, first counting only the best matched pairs, then counting top three Chinese translations for an English word. The top $N$ candidate evaluation is useful because in a machine-aided translation system, we could propose a list of up to, say, ten candidate translations to help the translator. We obtained the evaluations of three human judges (E1-E3). Evaluator E1 is a native Cantonese speaker, E2 a Mandarin speaker, and E3 a speaker of both languages. The results are shown in Figure 6.

The average accuracy for all evaluators for both sets is 73.1%. This is a considerable improvement from our previous algorithm (Fung & McKeown 1994) which found only 32 pairs of single word translation. Our program also runs much faster than other lexicon-based alignment methods.

We found that many of the mistaken translations resulted from insufficient data suggesting that we should use a larger size corpus in our future work. Tagging errors also caused some translation mistakes. English words with multiple senses also tend to be wrongly translated at least in part (e.g., *means*). There is no difference between capital letters and small letters in Chinese, and no difference between singular and plural forms of the same term. This also led to some error in the vector representation. The evaluators' knowledge of the language and familiarity with the domain also influenced the results.

Apart from single word to single word translation such as *Governor*/總督 and *prosperity*/繁榮, we also found many single word translations which show potential towards being translated as compound domain-specific terms such as follows:

- **finding Chinese words**: Chinese texts do not have word boundaries such as space in English, therefore our text was tokenized into words by a statistical Chinese tokenizer (Fung & Wu 1994). Tokenizer error caused some Chinese characters to be not grouped together as one word. Our program located some of these words. For example, *Green* was aligned to 綠, 皮 and 書 which suggests that 綠皮書 could be a single Chinese word. It indeed is the name for Green Paper – a government document.

- **compound noun translations**: *carbon* could be translated as 碳, and *monoxide* as 一氧化. If *carbon monoxide* were translated separately, we would get 碳 一氧化. However, our algorithm found both *carbon* and *monoxide* to be most likely translated to the single Chinese word 一氧化碳 which is the correct translation for *carbon monoxide*.

    The words *Legislative* and *Council* were both matched to 立法 and similarly we can deduce that Legislative Council is a compound noun/collocation. The interesting fact here is, *Council* is also matched to 局. So we can deduce that 立法局 should be a single Chinese word corresponding to *Legislative Council*.

- **slang**: Some word pairs seem unlikely to be translations of each other, such as *collusion* and its first three candidates 扯(it pull), 貓(cat), 尾(tail). Actually *pulling the cat's tail* is Cantonese slang for *collusion*.

    The word *gweilo* is not a conventional English word and cannot be found in any dictionary but it appeared eleven times in the text. It was matched to the Cantonese characters 俗, 稱, 鬼, and 佬 which separately mean *vulgar/folk*, *name/title*, *ghost* and *male*. 俗稱鬼佬 means *the colloquial term gweilo*. *Gweilo* in Cantonese is actually an idiom referring to a male westerner that originally had pejorative implications. This word reflects a certain cultural context and cannot be simply replaced by a word to word translation.

- **collocations**: Some word pairs such as *projects* and 房屋(*houses*) are not direct translations. However, they are found to be constituent words of collocations – the *Housing Projects* (by the Hong Kong Government). Both *Cross* and *Harbour* are translated to 海底(*sea bottom*), and then to 隧道(*tunnel*), not a very literal translation. Yet, the correct translation for 海底隧道 is indeed *the Cross Harbor Tunnel* and not *the Sea Bottom Tunnel*.

    The words *Hong* and *Kong* are both translated into 香港, indicating *Hong Kong* is a compound name.

    *Basic* and *Law* are both matched to 基本法, so we know the correct translation for 基本法 is *Basic Law* which is a compound noun.

- **proper names** In Hong Kong, there is a specific system for the transliteration of Chinese family names into English. Our algo-

| lexicons | total word pairs | correct pairs | | | accuracy | | |
| --- | --- | --- | --- | --- | --- | --- | --- |
| | | E1 | E2 | E3 | E1 | E2 | E3 |
| primary(1) | 128 | 101 | 107 | 90 | 78.9% | 83.6% | 70.3% |
| secondary(1) | 533 | 352 | 388 | 382 | 66.0% | 72.8% | 71.7% |
| total(1) | 661 | 453 | 495 | 472 | 68.5% | 74.9% | 71.4% |
| primary(3) | 128 | 112 | 101 | 99 | 87.5% | 78.9% | 77.3% |
| secondary(3) | 533 | 401 | 368 | 398 | 75.2% | 69.0% | 74.7% |
| total(3) | 661 | 513 | 469 | 497 | 77.6% | 71.0% | 75.2% |

Figure 4: Bilingual lexicon compilation results

rithm found a handful of these such as *Fung*/馮, *Wong*/黃, *Poon*/潘, *Hui*/ *Lam*/林, *Tam*/譚, etc.

## 7 Conclusion

Our algorithm bypasses the sentence alignment step to find a bilingual lexicon of nouns and proper nouns. Its output shows promise for compilation of domain-specific, technical and regional compounds terms. It has shown effectiveness in computing such a lexicon from texts with no sentence boundary information and with noise; fine-grain sentence alignment is not necessary for lexicon compilation as long as we have highly reliable anchor points. Compared to other word alignment algorithms, it does not need *a priori* information. Since EM-based word alignment algorithms using random initialization can fall into local maxima, our output can also be used to provide a better initializing basis for EM methods. It has also shown promise for finding noun phrases in English and Chinese, as well as finding new Chinese words which were not tokenized by a Chinese word tokenizer. We are currently working on identifying full noun phrases and compound words from noisy parallel corpora with statistical and linguistic information.